%% file: neurips_2019.tex
\documentclass{article}
\input{macros.tex}
\usepackage[dvipsnames]{xcolor}

\PassOptionsToPackage{square,numbers}{natbib}
     \usepackage[final]{neurips_2019}

\usepackage[utf8]{inputenc} 
\usepackage[T1]{fontenc}    
\usepackage{hyperref}       
\usepackage{url}            
\usepackage{booktabs}       
\usepackage{amsfonts}       
\usepackage{nicefrac}       
\usepackage{microtype}      

\usepackage[nottoc]{tocbibind}

\usepackage[square,numbers]{natbib}

\newtheorem{definition}{Definition}
\newtheorem{proposition}{Proposition}{}
\newtheorem{theorem}{Theorem}{}
\newtheorem{lemma}{Lemma}
\newtheorem{problem}{Problem}

\title{Abstraction based Output Range Analysis for Neural Networks}

\newcommand{\repeatthanks}{\textsuperscript{\thefootnote}}

\author{%
  Pavithra Prabhakar\thanks{Both authors contributed equally to this work.}  , \quad Zahra Rahimi Afzal\repeatthanks\\
  Department of Computer Science\\
  Kansas State University\\
  Manhattan, KS 66506\\
  \texttt{\{pprabhakar,zrahimi\}@ksu.edu} \\
}

\begin{document}

\maketitle
\input{abstract.tex} 

\input{intro.tex}
\input{inn.tex}

\input{abstraction.tex}

\input{encoding.tex}
\input{implementation}

\input{conclusion.tex}
\input{acknowledgment.tex}
\bibliographystyle{abbrvnat}
\bibliography{references}
\input{appendix.tex}


%
%
%
%
%
%

\end{document}

%% file: macros.tex
\usepackage{amssymb,amsmath,amsfonts,mathtools,amsthm}
\usepackage{enumerate}
\usepackage{flexisym}
\usepackage{graphicx}
\usepackage{cite}

\newcommand{\milp}{\ensuremath{\textit{MILP}}}
\newcommand{\I}{\ensuremath{\textit{I}}}
\newcommand{\enc}{\ensuremath{\textit{Enc}}}
\newcommand{\nn}{\ensuremath{\textit{NN}}}
\newcommand{\ab}[1]{\ensuremath{\textit{AV}(#1)}}
\newcommand{\val}{\ensuremath{\textit{Val}}}
\newcommand{\T}{\ensuremath{{\cal T}}}
\newcommand{\inn}{\ensuremath{\textit{INN}}}
\newcommand{\innt}{\ensuremath{\textit{INN T}}}

\newcommand{\relu}{\ensuremath{\textit{ReLU}}}
\newcommand{\norm}[1]{\ensuremath{|#1|}}

\newcommand{\real}{\ensuremath{\mathbb{R}}}

\newcommand{\tp}{\ensuremath{T/P}}
\newcommand{\df}[1]{\ensuremath{[|#1|]}}
\newcommand{\dfu}[2]{\ensuremath{\df{#1}_{#2}}}
\newcommand{\lwr}[1]{\ensuremath{#1^{l}}}
\newcommand{\uppr}[1]{\ensuremath{#1^{u}}}
\newcommand{\labsw}{\ensuremath{\lwr{\widehat{W}_i}}}
\newcommand{\labsb}{\ensuremath{\lwr{\hat{b}_i}}}
\newcommand{\uabsw}{\ensuremath{\uppr{\widehat{W}_i}}}
\newcommand{\uabsb}{\ensuremath{\uppr{\hat{b}_i}}}

\newcommand{\Si}{\ensuremath{\hat{S}_i}}
\newcommand{\Siplus}{\ensuremath{\hat{S}}_{i+1}}
\newcommand{\si}{\ensuremath{\hat{s}_i}} 
\newcommand{\siplus}{\ensuremath{\hat{s}}_{i+1}}

\newcommand{\h}[1]{\ensuremath{\hat{#1}}} 
\newcommand{\uabsbo}{\ensuremath{\uppr{\hat{b}_1}}}
\newcommand{\uabswz}{\ensuremath{\uppr{\widehat{W}_0}}}
\newcommand{\labswz}{\ensuremath{\lwr{\widehat{W}_0}}}
\newcommand{\labsbo}{\ensuremath{\lwr{\hat{b}_1}}}
\newcommand{\la}{\ensuremath {\textit{labs}}}
\newcommand{\ra}{\ensuremath {\textit{rabs}}}
\newcommand{\Post}{\ensuremath{\textit{Post}}}

%% file: abstract.tex
\begin{abstract}
In this paper, we consider the problem of output range analysis for
feed-forward neural networks with \relu{} activation functions.
The existing approaches reduce the output range analysis problem to
satisfiability and optimization solving, which are NP-hard problems,
and whose computational complexity increases with the number 
of neurons in the network. To tackle the computational complexity, we
present a novel abstraction technique that constructs a simpler neural
network with fewer neurons, albeit with interval weights called
interval neural network (\inn{}), which over-approximates the output
range of the given neural network. 
We reduce the output range analysis on the \inn{}s to solving a mixed
integer linear programming problem.
Our experimental results highlight the trade-off between the
computation time and the precision of the computed output range. 

\end{abstract}

%% file: intro.tex
\section{Introduction}
Neural networks are extensively used today in safety critical
control systems such as autonomous vehicles and airborne collision
avoidance systems~\citep{autocar1, autocar2, reluplex, aircraft1}.
Hence, rigorous methods to ensure correct functioning of neural
network controlled systems is imperative.
Formal verification refers to a broad class of techniques that provide
strong guarantees of correctness by exhibiting a proof.
Formal verification of neural networks has attracted a lot of attention
in the recent years~\citep{reluplex,bunel,range, taylor-nn1,
  taylor-nn2, taylor-nn3}.
However, verifying neural networks is extremely challenging due to the
large state-space, and the presence of nonlinear activation functions,
and the verification problem is known to be NP-hard for even simple
properties~\citep{reluplex}.

Our broad objective is to investigate techniques to verify neural
network controlled physical systems such as autonomous vehicles.
These systems consist of a physical system and a neural network
controller that are connected in a feedback, that is, the output of
the neural network is the control input (actuator values) to the
physical system and the output of the physical system (sensor values)
is input to the neural network controller.
An important verification problem is that of safety, wherein, one
seeks to ensure that the state of the neural network controlled system
never reaches an unsafe set of states.
This is established by computing the reachable set, the set of states
reached by the system, and ensuring that the reach set does not
intersect the unsafe states.
An important primitive towards computing the reachable set is to
compute the output range of a neural network controller given a set of
input valuations.

In this paper, we focus on neural networks with rectified linear unit
(\relu{}) function as an activation function, and we investigate the
output range computation problem for feed-forward neural
networks~\citep{range}.
Recently, there have been several efforts to address this problem
that rely on satisfiability checking and optimization.
Reluplex~\citep{reluplex} is a tool that develops a satisfiability
modulo theory for verifying neural networks, in  particular, it
encodes the input/output relations of a neural network as a
satisfiability checking problem. 
A mixed integer linear programming ($\milp$) based approach is
proposed in ~\citep{range, sriram2} to compute the output
range.
These approaches construct constraints that encode the neural network
behavior, and check satisfiability or compute optimal values over the
constraints.
The complexity of verification depends on the size of the constraints
which in turn depends on the number of neurons in the neural network.

To increase the verification efficiency, we present an orthogonal
approach that consists of a novel abstraction procedure to reduce the
state-space (number of neurons) of the neural network.
Abstraction is a formal verification technique that refers to methods for reducing the state-space while providing formal guarantees of properties that are preserved by the reduction. One of the well-studied abstraction procedures is predicate abstraction~\cite{clark11,graph} that consists of partitioning the state-space of a given system into a finite number of regions, and constructing an abstract system that consists of these regions as the states. Predicate abstraction has been employed extensively for safety verification, since, the safety of the abstract system is \emph{sound}, that is, it implies the safety of the given system. 
Our main result consists of a \emph{sound} abstraction that in
particular over-approximates the output range of a given neural
network.
Note that an over-approximation can still provide useful safety
analysis verdicts, since, if a superset of the reachable set does not
intersect with the unsafe set, then the actual reachable set will also
not intersect the unsafe set.
The abstraction procedure essentially merges sets of neurons within a
particular layer, and annotates the edges and  biases with interval
weights to account for the merging. 
Hence, we obtain a neural network with interval weights, which we call
\emph{interval neural networks} (\inn{}s). 
While interval neural networks are more general than neural networks,
we show as a proof of concept that the satisfiability and optimization
based verification approaches can be extended to \inn{}s by extending
the $\milp$ based encoding in~\citep{range} for neural networks to
interval neural networks and use it to compute the output range of the
abstract \inn{}.
We believe that other methods such as Reluplex can be extended to
handle interval neural network, and hence, the abstraction procedure
presented here can be used to reduce the state-space before applying 
existing or new verification algorithms for neural  networks.
An abstract interpretation based method has been explored in~\citep{rice}, wherein an abstract reachable set is propagated.
However, our approach has the flavor of predicate
abstraction~\citep{graph} and computes an over-approximate system
which can then be used to compute an over-approximation of the output
range using any of the above methods including the one based on
abstract interpretation~\citep{rice}.

The crucial part of the abstraction construction consists of
appropriately instantiating the weights of the abstract edges.
In particular, a convex hull of the weights associated with the
concrete edges corresponding to an abstract edge does not guarantee
soundness, which is shown using a counterexample in Section \ref{sec:abs}.
We need to multiply the convex hull by a factor equivalent to the
number of merged nodes in the source abstract node. 
The proof of soundness is rather involved, since, there is no
straightforward relation between the concrete and the abstract
states. 
We establish such a connection, by associating a set of abstract
valuations with a concrete valuation for a particular layer, wherein,
the abstract valuation for an abstract node takes values in the range
given by the concrete valuations for the related concrete nodes.
The crux of the proof lies in the observation (Proposition
\ref{prop:avg}) that the behavior of a concrete valuation is mimicked
in the abstract valuation by an average of the  concrete valuations at
the nodes corresponding to an abstract node. 
We conclude that the input/output relation associated with a certain
layer of the concrete system is over-approximated by input/output
valuations of the corresponding layer in the abstract system.

We have implemented our algorithm in a Python toolbox.
We perform experimental analysis on the ACAS~\citep{aircraft2} case
study, and observe that the verification time increases with the
increase in the number of abstract nodes, however, the
over-approximation in the output range decreases.
Further, we notice that the output range can vary non-trivially even
for a fixed number of abstract nodes, but different partitioning of
the concrete nodes for merging.
This suggests that further research needs to be done to understand
the best strategies for partitioning the state-space of neurons for
merging, which we intend to explore in the future. 
\paragraph{Related work.} 
Recent studies~\citep{taylor-survey,bunel,saftey1,taylor-nn3,saftey3,saftey4}
compare several neural  network verification algorithms.
Formal verification of feedforward neural networks with different
activation functions have been considered. For instance,
\citep{reluplex,rice} consider \relu{}, where as \citep{marta1,abstract1} consider large class of activation
 functions that can be represented as Lipschitz-continuous functions.
We focus on \relu{} functions, but our method can be extended to more
general functions.
Different verification problems have been considered including output
range analysis~\citep{shiram3,interval,marta1,saftey1,shiram4,taylor1}, and robustness analysis ~\citep{rice,Binarized}.
Verification methods include those based on reduction to
satisfiability solving~\citep{reluplex,saftey1,saftey3}, optimizaiton solving~\citep{anytime}, abstract interpretation~\citep{yasser,abstract1}, and linearization~\citep{saftey3,saftey4}.
There is some recent work on verification of AI controlled cyber-physical systems~\cite{new1,new2}

%% file: inn.tex
\section{Interval Neural Network}
\label{section.inn}
A neural network (\nn) is a computational model that
consists of nodes (neurons) that are organized in layers and edges
which are  the connections  between the nodes labeled by weights.
An $\nn$ contains an input layer, some hidden layers, and an output
layer each composed of neurons.
Given values to the nodes in the input layer, the values at the nodes
in the next layer are computed through a weighted sum dictated by the
edge weights and the addition of the bias associated with the output
node followed by an activation operation which we will assume is the $\relu$ 
(rectifier linear unit) function.
In this section, we introduce interval neural networks $\inn$ that
generalize neural networks with interval weights on edges and biases
and will represent our abstract systems.
\paragraph{Preliminaries.}
Let $\real$ denote the set of real numbers. Given a non-negative integer $k$, let $[k]$ 
denote the set $\{0,1, \cdots, k\}$. Given a set $A$, $\norm{A}$ represents the number of elements of $A$.
For any two functions $f, g: A \to \real$, we say $f\leq g$ if $\forall s\in A$, $f(s) \leq g(s)$. 
We denote the \relu{} function by $\sigma$, which is defined as ${\displaystyle \sigma(x) = \max(0,x)}$. Given two binary relations $R_1 \subseteq A \times B$ and $R_2 \subseteq B \times C$, we define their composition, denoted by
$R_1 \circ R_2$, to be $\{ (u, v) \ | \ \exists w, \ (u, w) \in R_1$ and $(w, v) \in R_2\}$. For any set $S$, a valuation over $S$ is a function $f : S \to \real$. We define $\val(S)$ to be the set of all valuations over $S$. 
A partition $R$ of the set $A$ is a set $R = \{R_1, \ldots, R_k\}$ such that 
$\bigcup_{i=1}^{k} R_i = A$
and  $R_i \cap R_j = \emptyset\quad  \forall i, j \in \{1, \ldots, k\}$ and $i \not= j$.

\begin{definition}[Interval Neural Network]\label{def:inn}
An interval neural network ($\inn$) is a tuple $(k, \{ S_i\}_{i \in
  [k]},$ $\{\lwr{W_i}, \uppr{W_i}\}_{i \in [k-1]}, \{\lwr{b_i},
\uppr{b_i}\}_{i \in [k]/\{0\}})$, where 
\begin{enumerate}[-]
\item $k$ is a natural number which we refer to as the number of layers;
\item $\forall i\in [k], \ S_i$ is a set of nodes of $i$-th layer in the
  interval neural network such that $\forall i \neq j, \ S_i \cap S_j
  = \emptyset$. $S_0$ is the input layer, $S_k$ is the output layer
  and $S_i, \ \forall i \in [k]/\{0,k\}$ is a hidden layer;
\item $\forall i\in [k-1], \ \lwr{W_i}, \uppr{W_i} : S_i\times S_{i+1}
  \to \real$ represent the weights of the edges between the $i$-th and
  $i+1$-th layer.
 We assume that $\forall i \in [k-1], s_i \in S_i, \ s_{i+1} \in S_{i+1}, \
  \lwr{W_i}(s_i,s_{i+1}) \leq \uppr{W_i}(s_i,s_{i+1})$;
\item $\forall i\in [k]/\{0\}, \ \lwr{b_i}, \uppr{b_i}: S_{i} \to
  \real$ are the biases associated with the nodes in the $i$-th layer,
  that is, $\forall i\in [k]/\{0\},
  s_{i} \in S_{i} , \ \lwr{b_i}(s_{i}) \leq \uppr{b_i}(s_{i})$.  
\end{enumerate}
\end{definition}




A neural network can be defined as a special kind of \inn{} where the
weights and biases are singular intervals. 
\begin{definition}[Neural Network]\label{def:nn}
An $\inn$ $\T$ is a neural network ($\nn$) if $\ \forall i \in [k-1],
s\in S_i, s\textprime \in S_{i+1}, \ \lwr{W_i}(s,s\textprime) =
\uppr{W_i}(s,s\textprime)$, and $\forall i\in [k]/\{0\}, s \in S_{i},
\ \lwr{b_i}(s) = \uppr{b_i}(s)$. 
\end{definition}
$\vspace{-0.1in}$
\begin{figure}[h]
\centering
\begin{minipage}{0.5\linewidth}
  \centering
  \includegraphics[width=\textwidth]{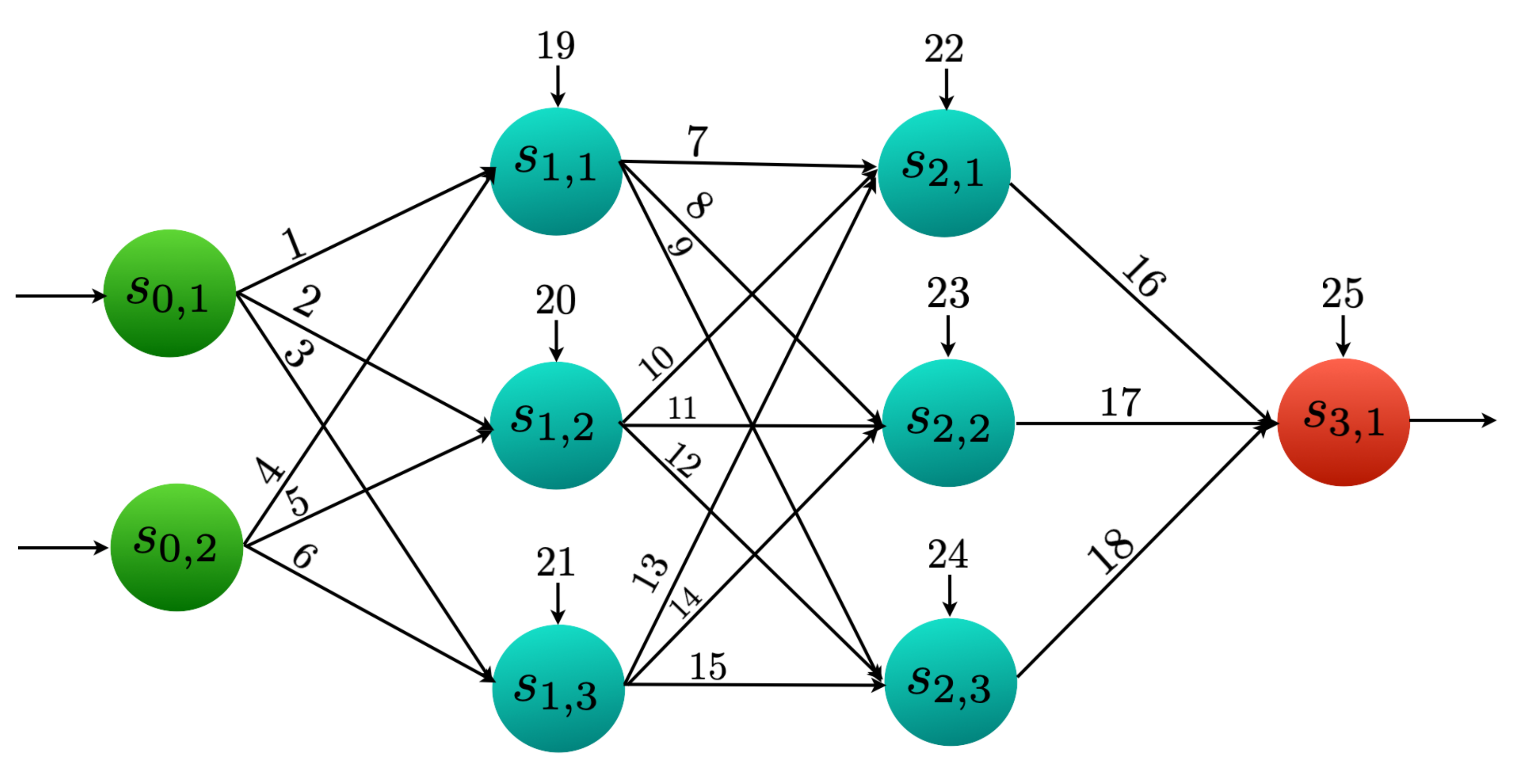}
  \caption{A neural network}
  \label{fig:nn}
\end{minipage}%
\begin{minipage}{0.5\linewidth}
  \centering
   \includegraphics[width=\textwidth]{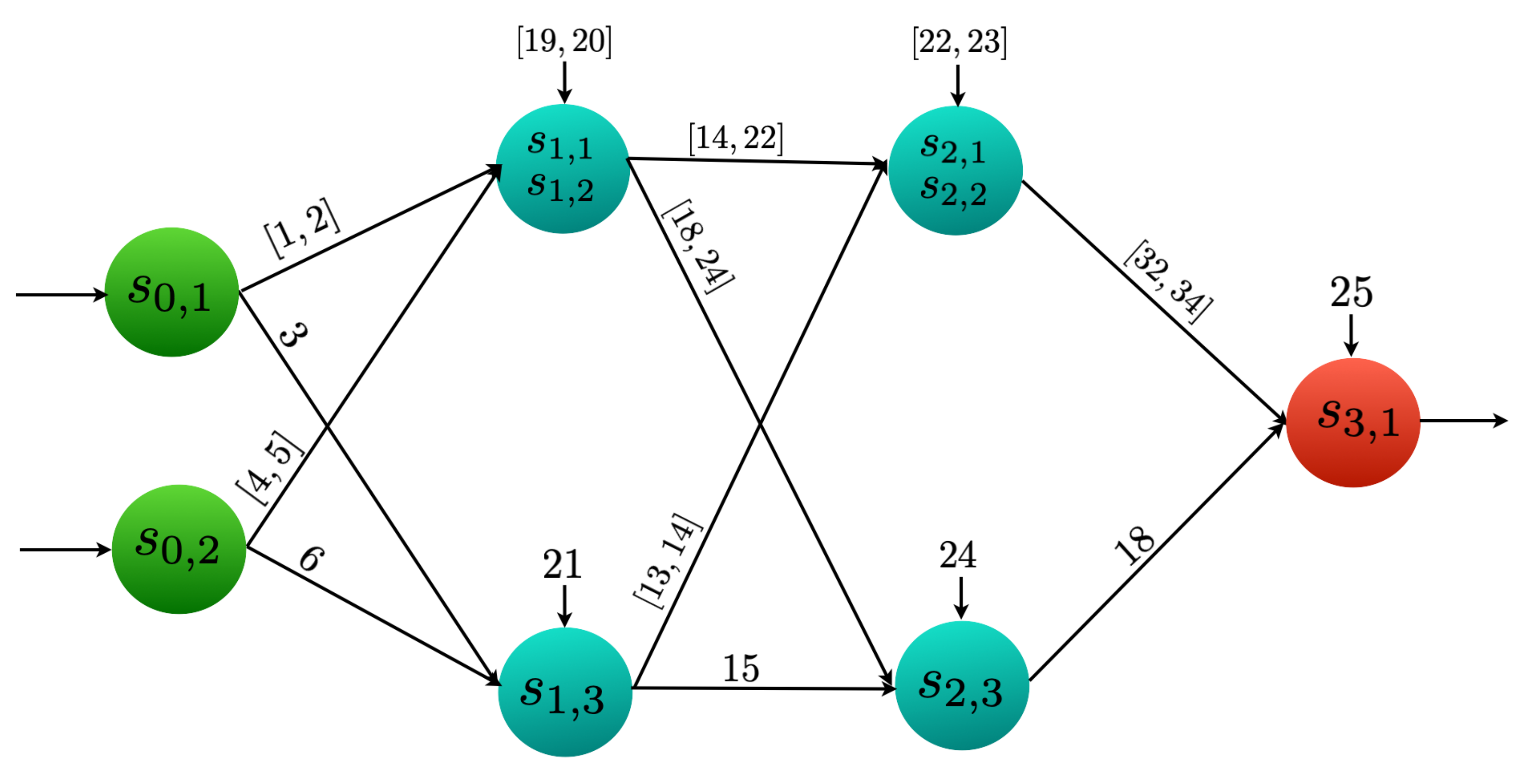}
  \caption{An interval neural network}
  \label{fig:inn}
\end{minipage}
\end{figure} 
Figure \ref{fig:nn} shows a neural network with $3$ layers. The input layer has $2$ nodes, the output layer has $1$ node, and each of the hidden layers has $3$ nodes.
The weights on the edges are a single number (singular intervals), hence, it is a neural network.
Figure \ref{fig:inn} shows an interval neural network again with $3$ layers.
The input and output layers have the same number of nodes as before, but the hidden layers have $2$ nodes each. 
The weights on the edges are intervals (and non-singular), so this is an interval neural network (rather than just a neural network).

An execution of the neural network starts with valuations to the input
nodes, and the valuations to the nodes of a certain layer are computed
based on the valuations for the nodes in the previous layer. 
More precisely, to compute the value at a node $s_{i, j}$ corresponding to the $j$-th node in the layer $i$, we choose a weight from the interval for each of the incoming nodes and compute a weighted sum of the valuations of the nodes in the previous layer. 
Then a bias is chosen from the bias interval associated with $s_{i, j}$ and added to the weighted sum. Finally, the $\relu$ function is applied on this sum.
The execution then proceeds to the next layer. 
The semantics of the neural network is captured using a set of pairs
of input-output valuations wherein the output valuation is a possible result starting from the input valuation and executing the neural network.
Next, we define the semantics of an $\inn$ as a set of valuations
for the input and output layers.
\begin{definition}[Semantics of \inn{}  Network]\label{def:sem_inn}
Given an $\inn$ $\T = (k, \{ S_i\}_{i \in  [k]},$ $\{\lwr{W_i},
\uppr{W_i}\}_{i \in [k-1]}, \{\lwr{b_i},\uppr{b_i}\}_{i \in [k]/\{0\}})$ and
$i \in [k-1]$, $\ \dfu{T}{i} = \{ (v_1,  v_2)\in \val(S_i) \times
\val(S_{i+1}) \, | \, $ $\forall s' \in S_{i+1}, \ v_2(s') = \sigma( \
\sum_{s \in S_i} w_{s,s'} \ v_1(s) + b_{s'}),$ where $\lwr{W_i}(s,
s') \leq w_{s,s'} \leq \uppr{W_i}(s, s'), \ \lwr{b_i}(s') \leq b_{s'}
\leq \uppr{b_i}(s')\}$.
We define $\df{T} = \dfu{T}{0} \circ \dfu{T}{1} \circ \cdots \circ \dfu{T}{k-1}$.
\end{definition}

The semantics can be captured alternately using a post operator, that
given a valuation of layer $i$, returns the set of all valuations of
layer $i+1$ that are consistent with the semantics.

\begin{definition}
Given an $\inn$ $\T$ with $k$ layers, $i \in [k-1]$ and $V \subseteq
\val(S_i)$, we define $\Post_{\T, i}(V) = \{v' \ |\ \exists v \in V,\
(v,v') \in \dfu{T}{i}\}$.
Given $V \subseteq \val(S_0)$, we define $\Post_\T(V) = \{v' \ |\
\exists v \in V,\ (v,v') \in \df{T}\}$.
\end{definition}
For notational convenience, we will write $\Post_{\T, i}(\{v\})$ and
$\Post_\T (\{v\})$ as just  $\Post_{\T, i}(v)$ and
$\Post_\T (v)$, respectively.


Our objective is to find an over-approximation of the
values the output neurons can take in an interval neural network, given
a set of valuations for the input layer.
\begin{problem}[Output range analysis]
Given an $\inn$ $\T$ with $k$ layers and a set of input valuations $I
\subseteq \val(S_0)$, compute valuations $l, u \in \val(S_k)$ such
that $\forall (v_1, v_2) \in \df{T}$  if $v_1 \in I$ then $l(s) \leq
v_2(s) \leq u(s)$ for every $s \in S_k$.
\end{problem}

%% file: abstraction.tex
\section{Our Approach}
\label{section.approach}
In this section, we present an abstraction based approach for
over-approximating the output range of an interval neural network.
First, in Section \ref{sec:abs}, we describe the construction of an abstract system whose semantics over-approximates the semantics of a given
$\inn$ and argue the correctness of the construction.
In Section \ref{sec:encode}, we present an encoding of the interval
neural network to mixed integer linear programming that enables the
computation of the output range. 

\subsection{Abstraction of an $\inn$}
\label{sec:abs}
\begin{figure}[h]
\centering
\begin{minipage}{0.5\linewidth}
  \centering
  \includegraphics[width=.7\textwidth]{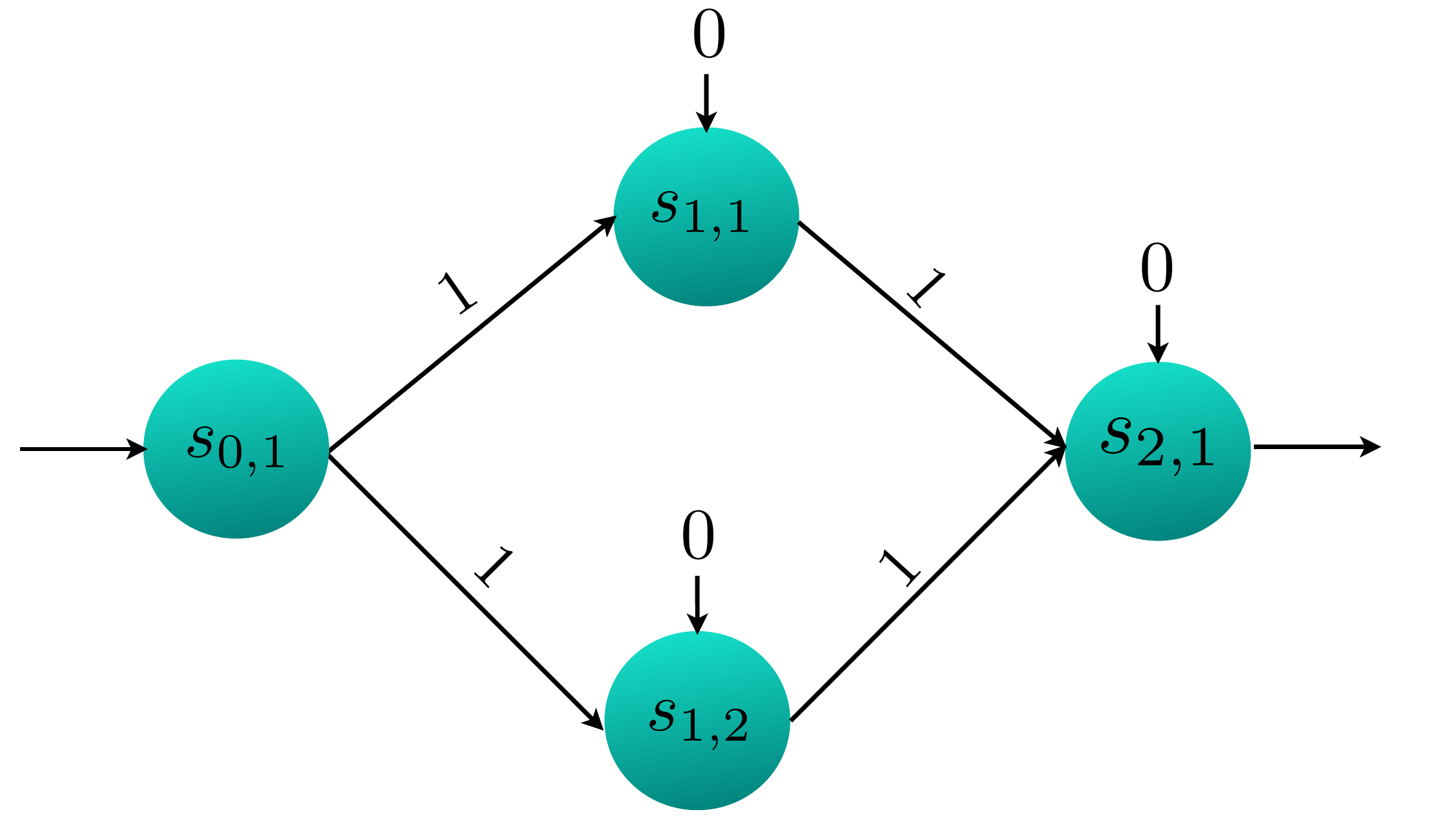}
  \caption{A concrete neural network}
  \label{fig:ex}
\end{minipage}%
\begin{minipage}{0.5\linewidth}
  \centering
   \includegraphics[width=.7\textwidth]{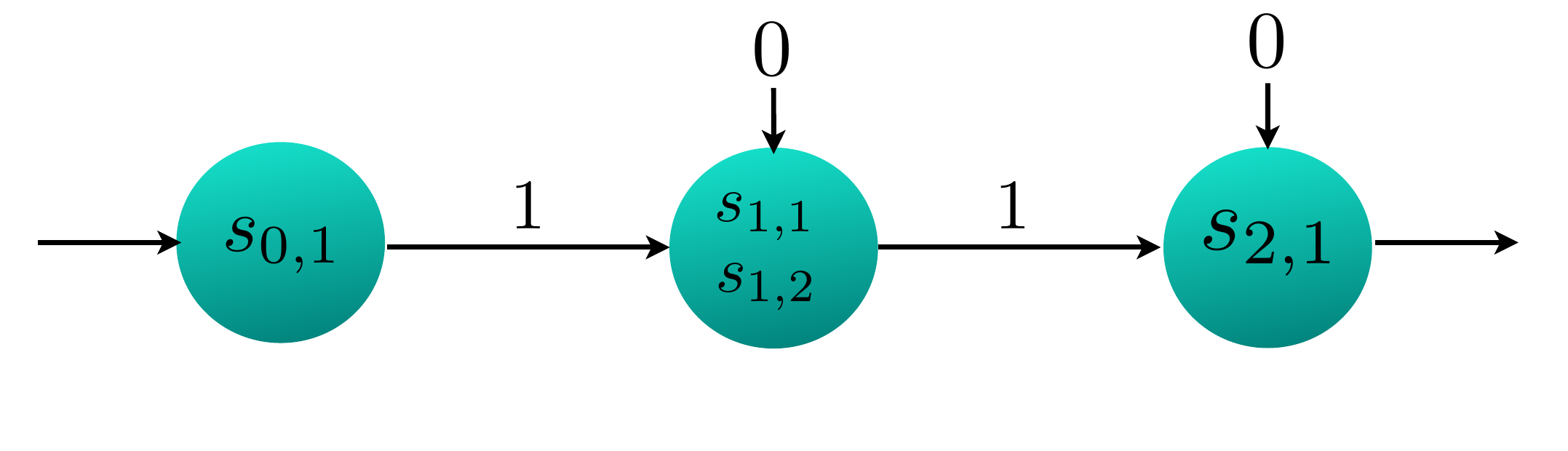}
  \caption{An incorrect abstraction}
  \label{fig:wrong}
\end{minipage}
\end{figure}

The motivation for the abstraction of an $\inn$ is to reduce the
``state-space'', the number of neurons in the network, so that
computation of the output range can scale to larger $\inn$s.
Our broad idea consists of merging the nodes of a given concrete
$\inn$ so as to construct a smaller abstract $\inn$.
However, it is crucial that we instantiate the weights on the edges
and the biases appropriately to ensure that the semantics of the
abstracted system is an over-approximation of the concrete $\inn$.
For instance, consider the neural network in Figure \ref{fig:ex} and
consider an input value $1$. It results in an output value of $2$.
Figure \ref{fig:wrong} abstracts the neural network in Figure
\ref{fig:ex} by taking the convex hull of the weights on the concrete
edges corresponding to the abstract edge. However, given input $1$,
the output of the abstract neural network is $1$ and does not contain
$2$.
Hence, we need to be careful in the construction of the abstract
system.
$\vspace{-0.1in}$

Given two sets of concrete nodes from consecutive layers of the $\inn$,
$\hat{s}_1$ and $\hat{s}_2$, which are each merged into one abstract
node, we associate an interval with the edge between $\hat{s}_1$ and
$\hat{s}_2$ to be the interval $ [|\hat{s}_1| w_1, |\hat{s}_1| w_2]$, where $w_1$ and $w_2$ are
the minimum and maximum weights associated with the edges in the
concrete system between nodes in $\hat{s}_1$ and $\hat{s}_2$,
respectively, and $|\hat{s}_1|$ is the number of concrete nodes
corresponding to the abstract node $\hat{s}_1$.
In other words, $[w_1, w_2]$ is the convex hull of the intervals
associated with the edges between nodes in $\hat{s}_1$ and
$\hat{s}_2$ multiplied by a factor corresponding to the number of
concrete nodes corresponding to the source abstract node.
Note that the above abstraction will lead to a weight of $2$ on the
second edge in Figure \ref{fig:wrong}, thus leading to an output of
$2$ as in the concrete system.

Next, we formally define the abstraction. We say that $P = \{P_i\}_{i
  \in [k]}$ is a partition of $\T$, if for every $i$, $P_i$ is a partition of the
$S_i$, the nodes in the $i$-th layer of $\T$. 
\begin{definition}[Abstract Neural Network]\label{def:abs}
Given an \innt{} $= (k, \{ S_i\}_{i \in [k]},$ $\{\lwr{W_i}, \uppr{W_i}\}_{i \in [k-1]}, \{\lwr{b_i}, \uppr{b_i}\}_{i \in [k]/\{0\}})$ and a partition $P = \{P_i\}_{i \in [k]}$ of $T$, we define an \inn{} $\tp = (k, \{ \Si\}_{i \in [k]}, \{\labsw, \uabsw\}_{i \in [k-1]}, \{\labsb, \uabsb\}_{i \in [k]/\{0\}})$, where 
\begin{enumerate}[-]
\item $\forall i\in [k], \ \Si = P_i$;
\item $\forall i\in [k-1]$, $\si \in \Si, \siplus \in \Siplus$, 
$\labsw(\si, \siplus) = |\h{s}_i|$ min $\{\lwr{W_i}(s_i, s_{i+1})\ | \
s_i\in \h{s}_i, \ s_{i+1}\in \h{s}_{i+1}\}$ and
$\uabsw(\si, \siplus) = |\h{s}_i|$  max $\{\uppr{W_i}(s_i, s_{i+1}) \ | \ s_i\in \h{s}_i, \ s_{i+1}\in \h{s}_{i+1}\}$;
\item $\forall i\in [k]/\{0\}, \si \in \Si$, 
$\labsb(\si) =$ min $\{\lwr{b_i}(s_{i})\ | \ s_{i}\in \h{s}_{i}\}$ and
$\uabsb(\si) =$ max $\{\uppr{b_i}(s_{i}) \ | \ s_{i}\in \h{s}_{i}\}$.
\end{enumerate}
\end{definition}

Figure \ref{fig:inn} shows the abstraction of the neural network in Figure \ref{fig:nn}, where the nodes $s_{1,1}$ and $s_{1,2}$ are merged and the nodes $s_{2,1}$ and 
$s_{2,2}$ are merged. 
Note that the edge from $\{s_{1, 1}, s_{1,2}\}$ to $\{s_{2,1}, s_{2,2}\}$ has weight interval $[14, 22]$, which is obtained by taking the convex hull of the four weights $7, 10, 8$ and $11$, and multiplying by $2$, the size of the source abstract node.

The following theorem states the correctness of the construction of
$\tp$. It states that every input/output valuation that is admitted by
$T$ is also admitted by $\tp$, thus establishing the soundness of the abstraction.
\begin{theorem}
  \label{thm:main}
Given an \innt{} $= (k, \{ S_i\}_{i \in [k]},$ $\{\lwr{W_i}, \uppr{W_i}\}_{i \in [k-1]}, \{\lwr{b_i}, \uppr{b_i}\}_{i \in [k]/\{0\}})$ and a partition $P = \{P_i\}_{i \in [k]}$ of $T$ such that $P_0 = S_0$ and $P_k = S_k$, $\df{T} \subseteq\df{\tp}$.   
\end{theorem}

We devote the rest of the section to sketch a proof of Theorem
\ref{thm:main}.
Broadly, the proof consists of relating the valuations in the $i$-th
layer of the concrete $\inn$ with the $i$-th layer of the abstract
$\inn$.
Note that the nodes in a particular layer of the abstract and the
concrete system might not be the same.
The following definition relates states in the concrete system to
those in the abstract system.

\begin{definition}
Given a valuation $v \in \val(S_i)$, $\ab{v} = \{\hat{v} \in
\val(\hat{S}_i) \,|\, \forall \hat{s} \in \hat{S}_i, \min_{s \in
  \hat{s}} v(s) \leq \hat{v}(\hat{s}) \leq \max_{s \in
  \hat{s}} v(s)\}$.
\end{definition}

Given a valuation $v$ of the $i$-th layer of the concrete system,
$\ab{v}$ consists of the set of all abstract valuations  of the $i$-th
layer in the abstract system, where each abstract node gets a value
which is within the range of values of the corresponding concrete
nodes.
Proof of Theorem \ref{thm:main} relies on the following connection
between corresponding layers of the concrete and abstract $\inn$s.

\begin{lemma}
  \label{lem:main}
If $(v,v')\in \dfu{T}{i}$, then $\ab{v'} \subseteq \Post_{\tp, i}(\ab{v})$. 
\end{lemma}

The proof of Lemma \ref{lem:main} broadly follows the following
structure.
We first observe that the abstraction procedure corresponding to edges
between layer $i$ and layer $i+1$ can be decomposed into
two steps, wherein we first merge the nodes of the $i$-th layer 
and then we merge the nodes of the $i+1$-st layer.
Note that
\[\labsw(\si, \siplus) = |\h{s}_i| \min_{ s_i\in \h{s}_i, \ s_{i+1}\in
    \h{s}_{i+1}} \lwr{W_i}(s_i, s_{i+1}) = \min_{\ s_{i+1}\in
    \h{s}_{i+1}} |\h{s}_i| \min_{ s_i\in \h{s}_i} \lwr{W_i}(s_i,
  s_{i+1})  \]
A similar observation can be made about the $\max$.
Hence, our first step consists of a function $\la$ which merges the
nodes in the ``left'' layer and associates an interval with the edges
which corresponds to computing the convex hull followed by multiplying
with an appropriate factor. Next, the function $\ra$ merges the
nodes in the ``right'' layer and associates an interval which
corresponds to only computing the convex hull.
Next, we define these abstraction functions, and state their relation
with the concrete systems.

\begin{definition}\label{def:labs}
Given an \inn{} $T= (k, \{ S_i\}_{i \in [k]}, \{\lwr{W_i},
\uppr{W_i}\}_{i \in [k-1]}, \{\lwr{b_i}, \uppr{b_i}\}_{i \in [k]/\{0\}})$,
$j$ and $P$ which is a partition of $j$th layer of $\T$, we define an
\inn{} $\la(\T, j, P) = (1, \{ \h{S_i} \}_{i \in [1]} , \{\hat{W}_i^l, \hat{W}_i^u\}_{i \in [0]}, \{\hat{b}_i^l, \hat{b}_i^u\}_{i \in \{1\}})$, where  
\begin{enumerate}[-]
\item $\h{S_0} = P, \ \h{S_1} = S_{j+1}$;
\item $\forall \h{s}_0 \in \h{S}_0, \h{s}_1 \in \h{S}_1$,
  $\labswz(\h{s}_0, \h{s}_1) =  |\h{s}_0|$ min $\{\lwr{W_0}(s_0, \h{s}_{1})\ | \
  s_0\in \h{s}_0\}$, and $\uabswz(\h{s}_0, \h{s}_1) = |\h{s}_0|$ max
  $\{\uppr{W}_0(s_0, \h{s}_{1}) \ | \ s_0\in \h{s}_0\}$; 
\item $\forall \h{s}_1 \in \h{S}_1,$ $\labsbo(\h{s}_1)
  =\lwr{b_1}(\h{s}_{1})$, and $\uabsbo(\h{s}_1)
  =\uppr{b_1}(\h{s}_{1})$.
\end{enumerate}
\end{definition}
$\vspace{-0.1in}$
\begin{figure}[h]
\centering
\begin{minipage}{0.5\linewidth}
  \centering
  \includegraphics[width=.4\textwidth]{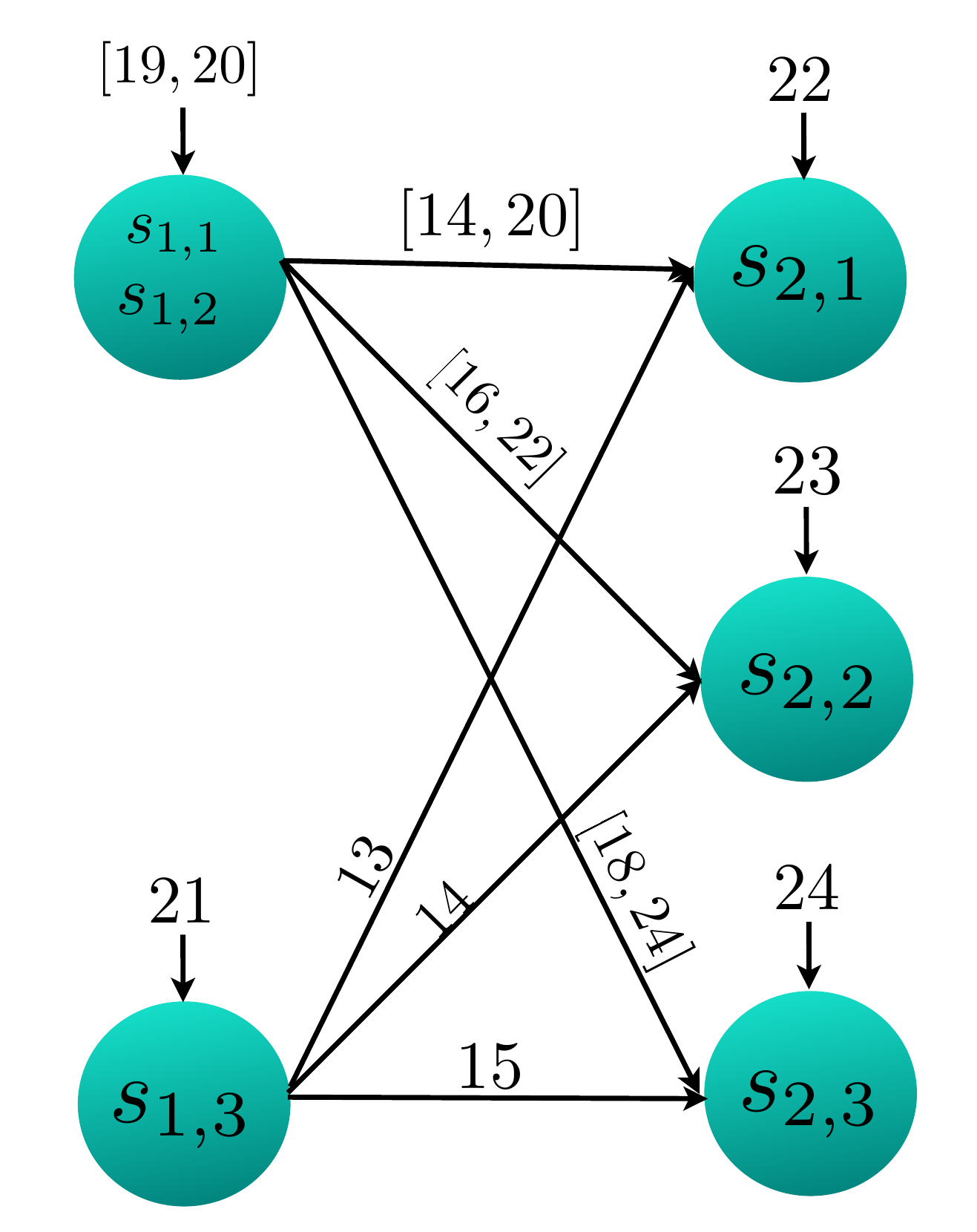}
  \caption{A left abstraction illustration}
  \label{fig:labs}
\end{minipage}%
\begin{minipage}{0.5\linewidth}
  \centering
   \includegraphics[width=.4\textwidth]{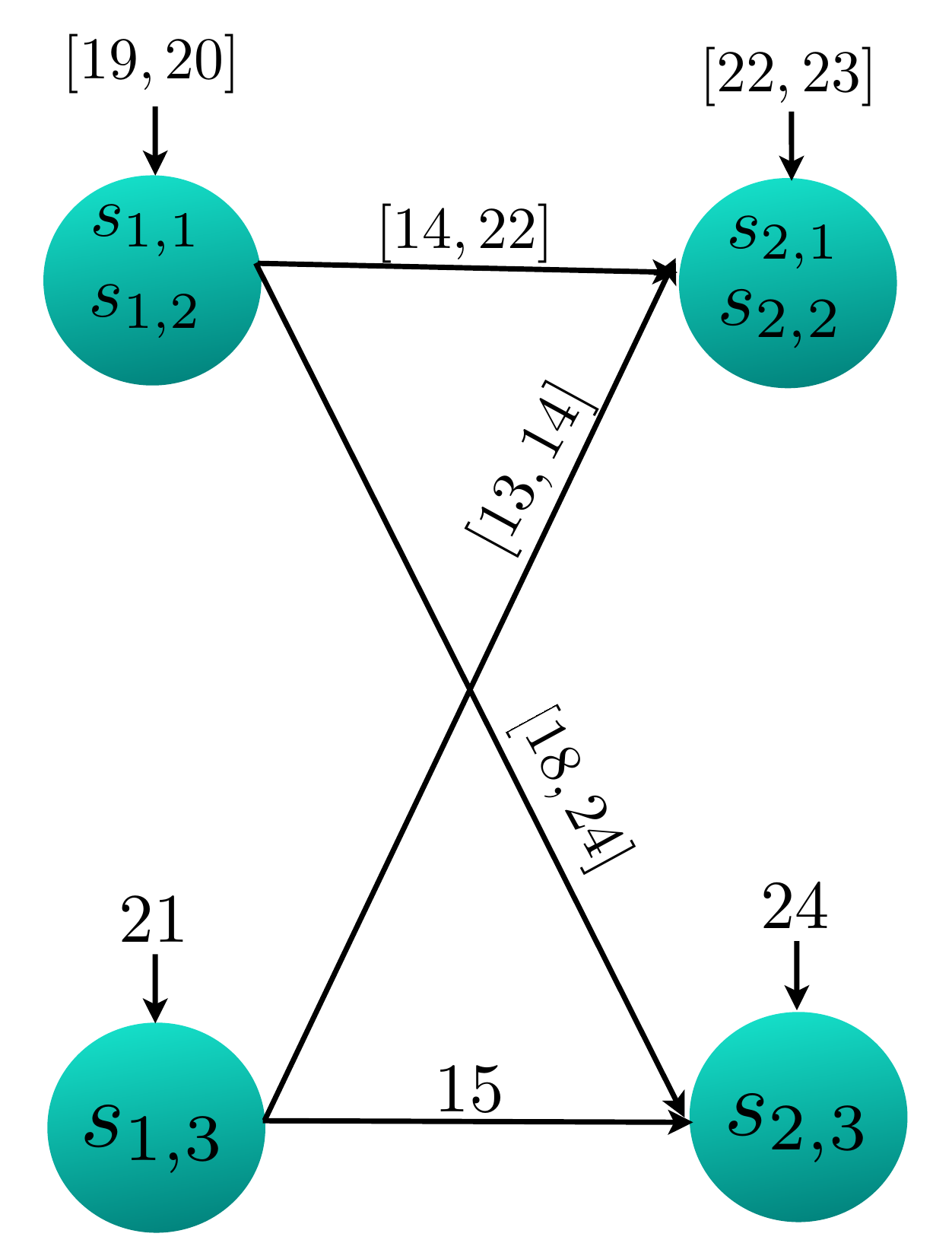}
  \caption{A right abstraction illustration}
  \label{fig:rabs}
\end{minipage}
\end{figure} 
Figure \ref{fig:labs} show the left abstraction of the the neural network in Figure \ref{fig:nn} with respect to layer $1$, where the nodes $s_{1,1}$ and $s_{1, 2}$ are merged. The edge from $\{s_{1,1}, s_{1,2}\}$ to $s_{2, 1}$ has weight $[14, 20]$ which is obtained by taking the convex hull of the values $7$ and $10$ and multiplying by $2$.

\begin{definition}{}
\label{def:rabs}
Given an \inn{} $\T=(1, \{ S_i\}_{i \in [1]}, \{\lwr{W_i},
\uppr{W_i}\}_{i \in [0]} \},
\{\lwr{b_i}, \uppr{b_i}\}_{i \in \{1\}})$ and $P$ which is a partition of the 
layer $1$ of $\T$, we define an \inn{} $\ra(\T,P) = (1, \{ \h{S_i} \}_{i \in [1]} , \{\hat{W}_i^l, \hat{W}_i^u\}_{i \in [0]}, \{\hat{b}_i^l, \hat{b}_i^u\}_{i \in \{1\}})$, where  
\begin{enumerate}[-]
\item $\h{S_0} = S_{0}, \ \h{S_1} = P$;
\item $\forall \h{s}_0 \in \h{S}_0, \h{s}_1 \in \h{S}_1$,
  $\labswz(\h{s}_0, \h{s}_1) = $ min $\{\lwr{W_0}(\h{s}_0, s_{1})\ | \
  s_1\in \h{s}_1\}$, and $\uabswz(\h{s}_0, \h{s}_1) =$ max
  $\{\uppr{W_0}(\h{s}_0, s_{1}) \ | \ s_1\in \h{s}_1\}$; 
\item $\forall \h{s}_1 \in \h{S}_1, \ \labsbo(\h{s}_1) =$ min $\{
  \lwr{b_1}(s_{1})\ | \ s_{1}\in \h{s}_1 \}$, and $\uabsbo(\h{s}_1) =$
  max $\{ \uppr{b_1}(s_{1})\ | \ s_{1}\in \h{s}_1 \}$. 
\end{enumerate} 
\end{definition}
Figure \ref{fig:rabs} shows the right abstraction of the interval neural network in Figure \ref{fig:labs}, where the nodes $s_{2,1}$ and $s_{2, 2}$ are merged. The edge from $\{s_{1,1}, s_{1,2}\}$ to $\{s_{2, 1}, s_{2, 2}\}$ has weight $[14, 22]$ which is obtained by taking the convex hull of the intervals $[14, 20]$ and $[16, 22]$. Note that Figure \ref{fig:rabs} is the same as the Figure \ref{fig:inn} with restricted $2$ layers $1$ and $2$.

Note that applying the left abstraction followed by right abstraction
to the $j$-th layer of $\T$ gives us the $j$-th layer of $\tp$. This is
stated in the following lemma.

\begin{lemma}{}
  \label{lem:rl}
 $\ra(\la(\T, j, P_j), P_{j+1}) = (1, \{ \hat{S}_{i+j}\}_{i \in [1]},
 \{\hat{W}_{i+j}^l, \hat{W}_{i+j}^u\}_{i \in [0]}, \{\hat{b}_{i+j}^l,$ $
 \hat{b}_{i+j}^u\}_{i \in \{1\}})$. 
  \end{lemma}

Proof of Lemma \ref{lem:main} relies on some crucial properties which
we state below.
The crux of the proof of the correctness of left abstraction lies in
the following proposition. It states that the contribution of the
values of a set of left nodes $\{s_1, \cdots, s_n\}$ on a right node $s$, can be
simulated in a left abstraction which merges $\{s_1, \cdots, s_n\}$ by
the average of the values.

\begin{proposition}
  \label{prop:avg}
Let $v_1, v_2, \ldots, v_n$ and $w_1, w_2, \ldots, w_n$ be real
numbers.
Let $\bar{v} = \sum_i v_i / n$.
There exists a $w$ such that $n \min_i w_i \leq w \leq n \max_i w_i$ and
$\sum_i w_i v_i = \bar{v} w$.
\end{proposition}

\begin{proposition}
  \label{prop:labs}
If $(v_1,v_2)\in\dfu{T}{i}$, then $v_2\in \Post_{\la(\T, j, P)}(\ab{v_1})$.
\end{proposition}

Next, we state the correctness of $\ra$. Here we show that
given any valuation $v$ of the right layer in the concrete system, any
valuation $\hat{v} \in \ab{v}$ can be obtained in the abstraction.
It relies on the observation that $\hat{v}(\hat{s})$ is a convex
combination of the $\min_{s \in \hat{s}} v(s)$ and $\min_{s \in
  \hat{s}} v(s)$, and the weight interval of an abstract edge is a convex hull
of the intervals of the corresponding concrete edges.

\begin{proposition}
 \label{prop:rabs}
Given an $\inn$ $\T$ with one layer, and a partition $P$ of layer $1$, if $(v_1,v_2)\in\df{T}$, then
$\ab{v_2} \subseteq \Post_{\ra(\T, P)}(v_1)$.
\end{proposition}


Proofs are eliminated due to shortage of space and are provided in the
supplementary material.

%% file: encoding.tex
\subsection{Encoding the interval neural network and \milp{} solver}
\label{sec:encode}
In this section, we present a reduction of the range computation
problem to solving a mixed integer linear program. 
The ideas are similar to those in~\citep{saftey5, cheng} using the big-M method.
However, since, our weights on the edges are not unique but come from
an interval, a direct application of the previous encodings where the
constant weights are replaced by a variable with additional
constraints related to the interval the weight variable is required to
lie in, results in non-linear constraints.
However, we observe that we can eliminate the weight variable by
replacing it appropriately with the minimum and maximum values of the
interval corresponding to it.

We encode the semantics of an $\inn$ $\T$ as a constraint $\enc(\T)$
over the following variables. For every node $s$ of the $\inn$ $\T$, we have a real valued variable $x_s$, and we have a binary variable $q_s$ that takes values in $\{0, 1\}$.
Let $X_i$ denote the set of variables $\{x_s \,|\, s \in S_i\}$, and
$Q_i = \{q_s \ | \ s \in S_i\}$.
Given a valuation $v \in \val(S_i)$, we will abuse notation and use
$v$ to also denote a valuation of $X_i$, wherein $v$ assigns to $x_s
\in X_i$, the valuation $v(s)$, and vice versa.
Let $X = \cup_i X_i$ and $Q = \cup_i Q_i$.
$\enc(\T)$ is the union of the encodings of the different layers of
$\T$, that is, $\enc(\T) = \cup_i \enc(\T, i)$, where $\enc(\T, i)$
denotes the constraints corresponding to layer $i$ of $\T$.
$\enc(\T, i)$ in turn is the union of constraints corresponding to the
different nodes in layer $i+1$, that is, $\enc(\T, i) = \cup_{s' \in
  S_{i+1}} C_{s'}^{i+1}$, where the constraints in $C_{s'}^{i+1}$ are
as below:
\begin{align}
C_{s'}^{i+1}:    
\begin{cases} 
      \sum_{s \in S_i} \lwr{W_i}(s, s')x_s +\lwr{b_{i}}(s') \leq x_{s'}, \ 0\leq x_{s'}\\
       \sum_{s \in S_i} \uppr{W_i}(s, s')x_s +\uppr{b_{i}}(s')+ Mq_{s'} \geq x_{s'}, \ M(1-q_{s'}) \geq x_{s'}
\end{cases}
\end{align}
Here, $M$ is an upper bound on the absolute values any neuron can take (before applying the $\relu$ operation) for a given input set. 
It can be estimated using the norms
of the weights interpreted as matrices, that is, $\lVert \lwr{W_i}
\rVert$ and $\lVert \uppr{W_i}
\rVert$, an interval box around the input polyhedron, and the norms of
biases.

Next, we state and prove the correctness of the encoding. More
precisely, we show that $(v, v') \in \dfu{T}{i}$ if and only if there are
valuations for the variables in $Q_i$ such that the constraints
$\enc(T, i)$ are satisfied when the values for $X_i$ and $X_{i+1}$ are
provided by $v$ and $v'$.

\begin{theorem}\label{theorm:encoding}
  Let $v \in \val(S_i)$ and $v' \in \val(S_i)$.
Then $(v, v') \in \dfu{T}{i}$ if and only if there is a valuation $z
\in \val(Q_i)$, such that $\enc(T, i)$ is satisfied with values $v,
v'$ and $z$.
\end{theorem} 
We can now compute the output range analysis by solving a maximization
and a minimization problem for each output variable.
More precisely, for each $s \in S_k$, the output layer, we solve:
$\max x_s$ such that $\enc(\T)$ and $\I$ hold, where $\I$ is a
constraint on the input variables encoding the set of input
valuations.
Similarly, we solve a minimization problem, and thus obtain an output
range for the variable $x_s$ given the input set of valuations $\I$.
The maximization and minimization problems can be solved using mixed
integer linear programming ($\milp$) if $\I$ is specified using linear
constraints.
Even checking satisfiability of a set of mixed integer linear
constraints is NP-hard problems, however, there are commercial software tools
that solve $\milp$ such as Gurobi and CPLEX.

%% file: implementation.tex
\section{Implementation}
In this section, we present our experimental analysis using a Python
toolbox that implements the abstraction procedure and the reduction of
the $\inn$ output range computation to $\milp$ solving.
We consider as a case study ACAS Xu benchmarks, which are neural
networks with $6$ hidden layer with each layer consisting of $50$
neurons~\citep{bunel}.
We report here the results with one of the benchmarks, we observed
similar behavior with several other benchmarks.

\begin{figure}[h]
\centering
\begin{minipage}{0.5\textwidth}
  \centering
  \includegraphics[width=.7\textwidth]{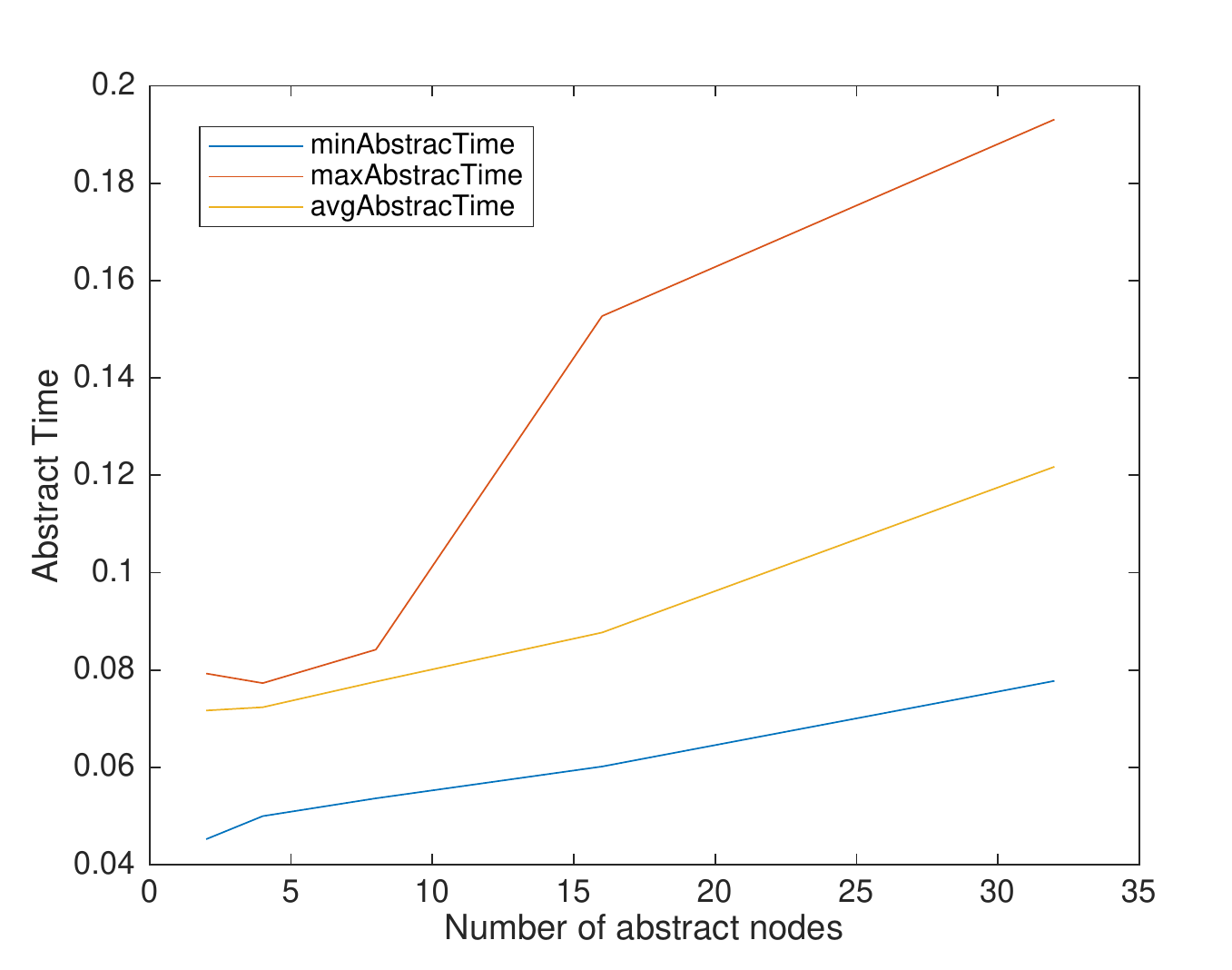}
  \caption{Abstraction Time}
  \label{fig:abstract}
\end{minipage}%
\begin{minipage}{0.5\textwidth}
  \centering
   \includegraphics[width=.7\textwidth]{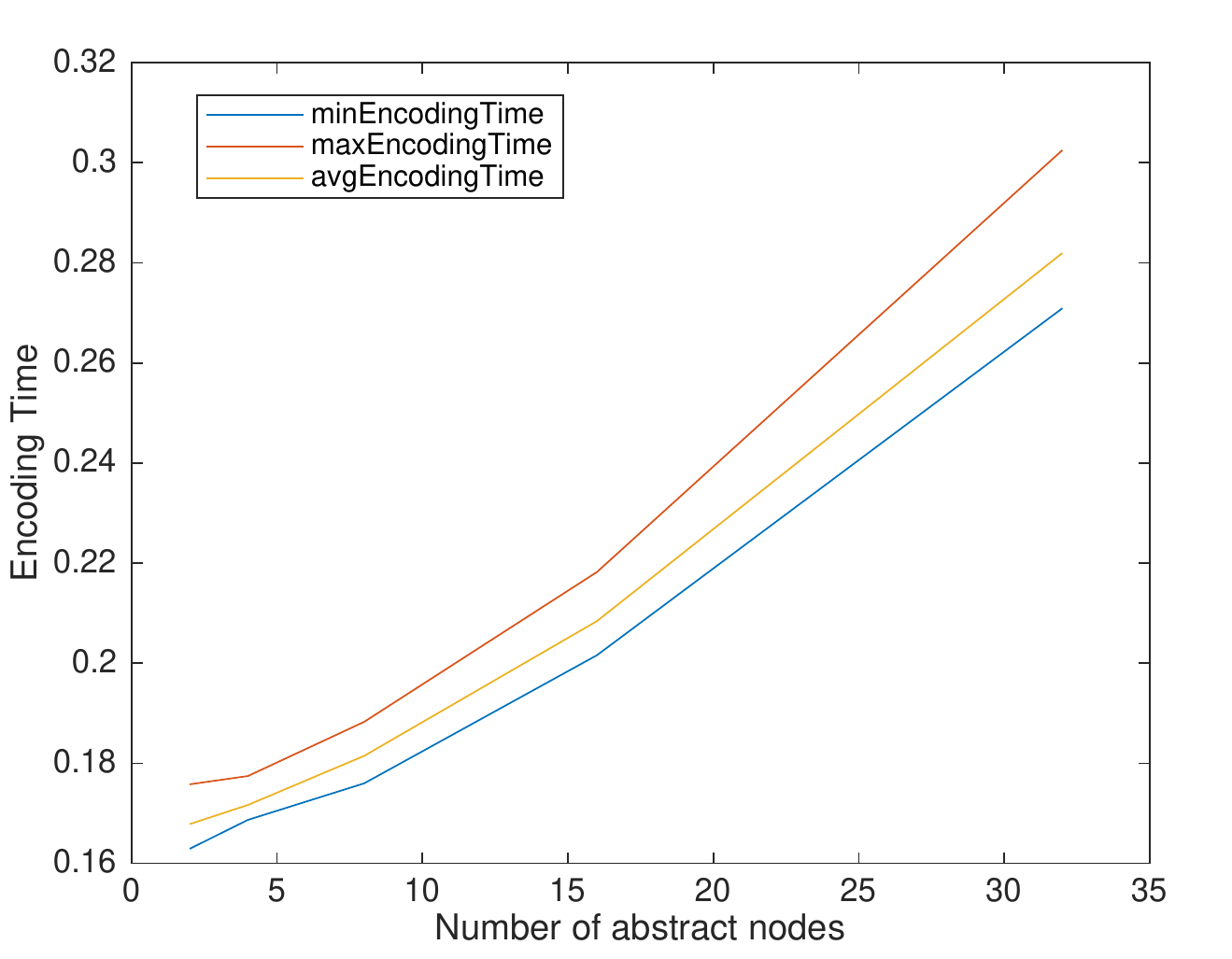} 
  \caption{Encoding Time}
  \label{fig:encoding}
\end{minipage}
\end{figure}

\begin{figure}[h]
\centering
\begin{minipage}{0.5\textwidth}
  \centering
  \includegraphics[width=.7\textwidth]{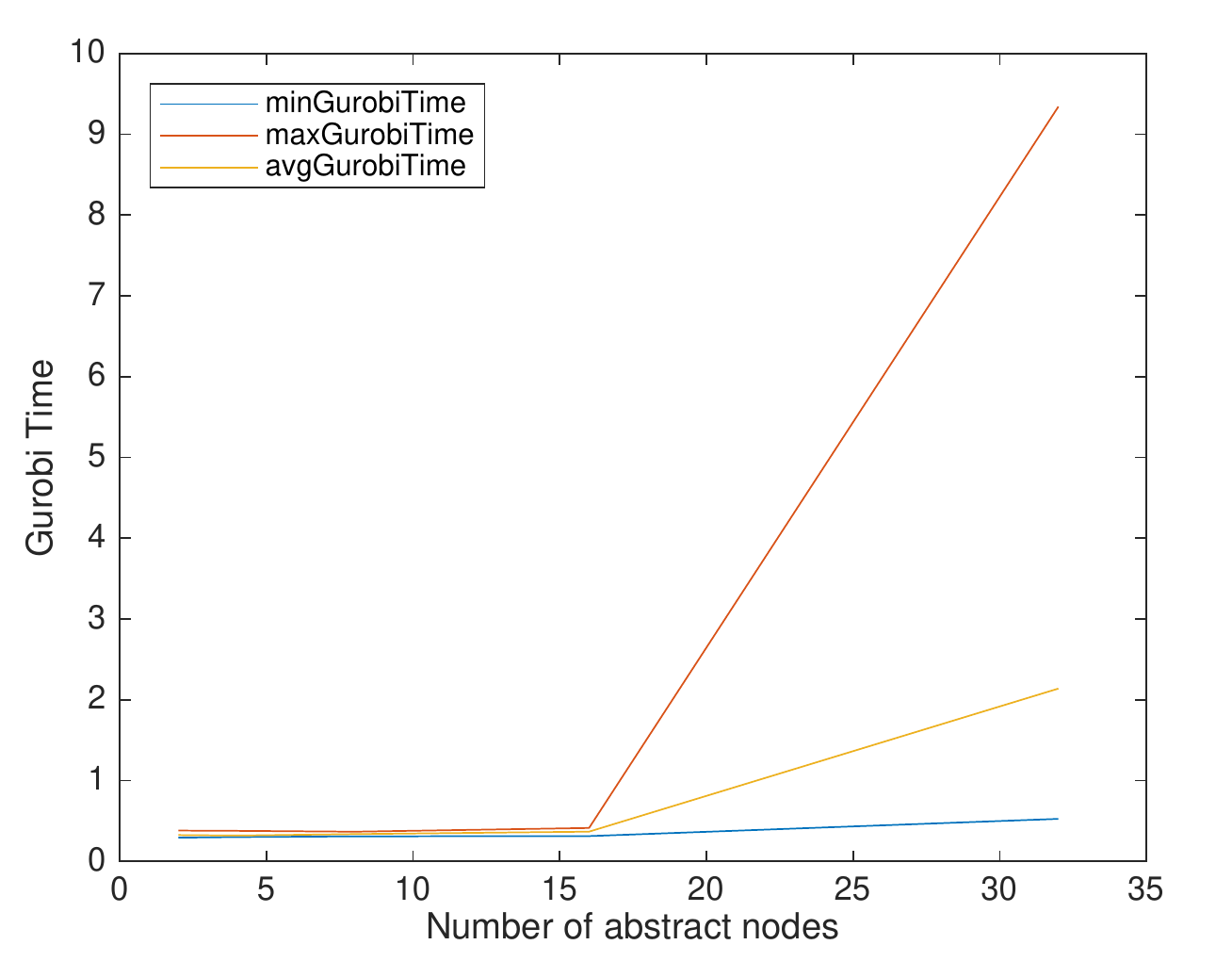}
  \caption{\milp{} Solving Time}
  \label{fig:gurobi}
\end{minipage}%
\begin{minipage}{0.5\textwidth}
  \centering
   \includegraphics[width=.7\textwidth]{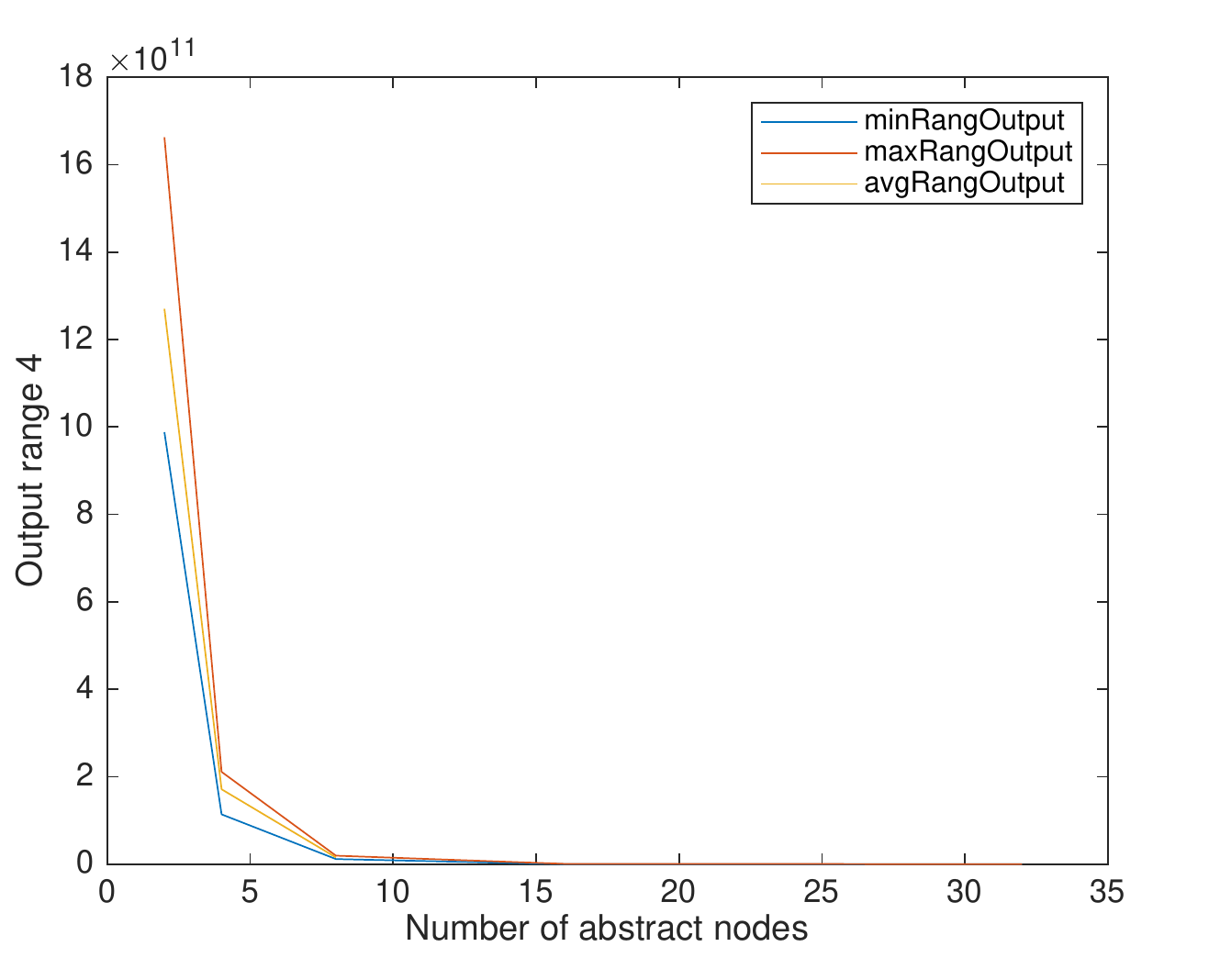}
  \caption{Output Range}
  \label{fig:range}
\end{minipage}
\end{figure}

We consider abstractions of the benchmark with different number of
abstract nodes, namely, $2, 4, 8, 16, 32$, which are generated
randomly.
For a fixed number of abstract nodes, we perform $30$ different random
runs, and measure the average, maximum and minimum time for different
parts of the analysis.
Similarly, we compute the output range for a fixed number of abstract
nodes, and obtain the average, maximum and minimum on the lower and
upper bound of the output ranges.  The lower bound was unanimously
$0$, hence, we do not report it here.
The results are summarized in Figures
~\ref{fig:abstract},~\ref{fig:encoding},~\ref{fig:gurobi}, and~\ref{fig:range}.

As shown in Figure~\ref{fig:abstract}, the abstraction construction
time increases gradually with the number of abstract neurons.
We observe a similar trend with encoding time.
However, the time taken by Gurobi to solve the $\milp$ problems
increases drastically after certain number of abstract nodes. Also, as
shown in Figure~\ref{fig:gurobi}, the $\milp$ solving time by Gurobi
is the most expensive part of the overall computation.
Since, this is directly proportional to the number of abstract nodes,
abstraction procedure proposed in the paper, has the potential to
reduce the range computation time drastically.
In fact, Gurobi did not return when ACAS Xu benchmark was encoded
without any abstraction, thus, demonstrating the usefulness of the
abstraction.
$\vspace{-0.1in}$

We compare output ranges (upper bounds) based on different
abstractions. The upper bound of the output range decreases as we consider
more abstract nodes, since, the system becomes more precise.
In fact, it decreases very drastically in the first few abstraction.
We compute the average, minimum and maximum of the upper bound on the
output range.
Even for a fixed number of abstract nodes, the maximum and minimum of
the upper bound on the output range among the random runs has a wide
range, and depends on the specific partitioning.
For instance, as seen in the Figure~\ref{fig:range}, although we have
only $2$ partitions, the upper bound on the output range varies by a
factor of $2$. 
This suggest that the partitioning strategy can play a crucial role in
the precision of output range.
Hence, we plan to explore partitioning strategies in the future.
To conclude, our method provides a trade-off between
verification time and the precision of the output range depending on
the size of the abstraction.

%% file: conclusion.tex
\section{Conclusions}
In this paper, we investigated a novel abstraction techniques for
reducing the state-space of neural networks by introducing the concept
of interval neural networks. Our abstraction technique is orthogonal
to existing techniques for analyzing neural networks.
Our experimental results demonstrate the usefulness of abstraction
procedure in computing the output range of the neural network, and the
trade-off between the precision of the output range and the computation
time.
However, the precision of the output range is affected by the specific
choice of the partition of the concrete nodes even for a fixed number
of abstract nodes. Our future direction will consist of exploring
different partition strategies for the abstraction with the aim of obtaining
precise output ranges. In addition, we will consider more complex
activation function. Our abstraction technique will extend in a
straightforward manner, however, we will need to investigate methods
for analyzing the ``interval'' version of the neural network for these
new activation functions.

%% file: acknowledgment.tex
\section*{Acknowledgments}
Pavithra Prabhakar was partially supported by NSF CAREER Award No. 1552668 and  ONR YIP Award No. N000141712577.

%% file: appendix.tex
\section{Supplementary material}
\paragraph{Proof of Proposition \ref{prop:avg}.}
Let $w = \sum_i w_i v_i / \bar{v}$. Then it trivially satisfies
$\sum_i w_i v_i = \bar{v} w$.
Let $w_\min = \min_i w_i$ and $w_\max = \max_i w_i$.
We need to show that $n w_\min \leq w \leq n w_\max$.
Note that $w = \sum_i w_i v_i / \bar{v}  \geq \sum_i w_\min v_i /
\bar{v} = w_\min \sum_i v_i /
\bar{v} = w_\min n \bar{v}/\bar{v}  = n w_\min$.
Similarly, we can show that $w = \sum_i w_i v_i / \bar{v}  \leq n w_\max$.
The following proposition captures the relation between a layer of the
concrete system and its left abstraction.

\paragraph{Proof of Proposition \ref{prop:labs}.}
  Let $(v_1,v_2)\in\dfu{T}{i}$.
  Let $ S_0$ and $S_{1}$ be the left and right layers of
  $\la(\T, j,$ $ P)$, respectively.
From the definition of the semantic of \inn{} given by Definition
\ref{def:sem_inn}, we know that for any $s' \in S_1$,  $v_2(s') =
\sigma(\sum_{s \in S_0} w_{s, s'} v_1(s) + b_{s'})$, where for every
$s$, $\lwr{W_0}(s, s') \leq w_{s,s'} \leq \uppr{W_0}(s, s'), \
\lwr{b_1}(s') \leq b_{s'} \leq \uppr{b_1}(s')$.
We can group together all neurons that are merged together in $P$, and
rewrite the above as
$v_2(s') = \sigma(\sum_{\hat{s} \in P} \sum_{s \in \hat{s}} w_{s, s'}
v_1(s) + b_{s'})$.
From Proposition \ref{prop:avg}, we can replace $\sum_{s \in \hat{s}} w_{s, s'}
v_1(s)$ by $v_{\hat{s}} w_{\hat{s}}$, where $v_{\hat{s}} = \sum_{s \in
  \hat{s}} v_1(s) / \norm{\hat{s}}$ and $w_{\hat{s}}$ is such that
$\norm{\hat{s}} \min_{s \in \hat{s}} w_{s, s'} \leq w_{\hat{s}} \leq
\norm{\hat{s}} \max_{s \in \hat{s}} w_{s, s'}$.
Consider a valuation $\hat{v}_1$, where $\hat{v}_1(\hat{s}) =
v_{\hat{s}} = \sum_{s \in
  \hat{s}} v_1(s) / \norm{\hat{s}}$. Since, the average is in between the minimum and maximum
values, $\hat{v}_1 \in \ab{v_1}$.
Now $v_2$ can be rewritten using $\hat{v}_1$ as
$v_2(s') = \sigma(\sum_{\hat{s} \in P} v_{\hat{s}} w_{\hat{s}} +
b_{s'}) = \sigma(\sum_{\hat{s} \in P} \hat{v}_1(\hat{s})  w_{\hat{s}} +
b_{s'})$, where $\norm{\hat{s}} \min_{s \in \hat{s}} w_{s, s'} \leq w_{\hat{s}} \leq
\norm{\hat{s}} \max_{s \in \hat{s}} w_{s, s'}$.
Since $w_{s, s'}$ also satisfies $\norm{\hat{s}} \min_{s \in \hat{s}}
W_0^l(s, s') \leq w_{\hat{s}} \leq
\norm{\hat{s}} \max_{s \in \hat{s}}$ $ W_0^u(s, s')$, we see that
$v_2\in \Post_{\la(\T, j, P)}(\ab{v_1})$ (since, $v_2$ and
$\hat{v}_1$ satisfy the semantics of $\la(\T, j, P)$).

\paragraph{Proof of Proposition \ref{prop:rabs}.}
Consider $\hat{v}_2 \in \ab{v_2}$. Then $\hat{v}_2({\hat{s}}') = \alpha
v_2(s'_1) + (1 - \alpha) v_2(s'_2)$, where $s'_1$ is the node in
${\hat{s}}'$ for which $v_2$ at the node is the minimum and $s'_2$ is the node in
${\hat{s}}'$ for which $v_2$ at the node is the maximum.
Let $S_0$ and $S_1$ be the nodes in the left and right layers of $\T$.
$v_2(s'_i) = \sigma(\sum_{s \in S_0} w_{s, s'_i} v_1(s) + b_{s'_i})$, where for every
$s$, $\lwr{W_0}(s, s'_i) \leq w_{s,s'_i} \leq \uppr{W_0}(s, s'_i), \
\lwr{b_1}(s'_i) \leq b_{s'_i} \leq \uppr{b_1}(s'_i)$.
$\hat{v}_2({\hat{s}}') = \alpha
v_2(s'_1) + (1 - \alpha) v_2(s'_2) = \alpha \sigma(\sum_{s \in S_0}
w_{s, s'_1} v_1(s) + b_{s'_1}) + (1 - \alpha) \sigma(\sum_{s \in S_0}
w_{s, s'_2} v_1(s) + b_{s'_2})$.
Let us first consider the case where the expressions within $\sigma$
are non-negative.
Then $\hat{v}_2({\hat{s}}') = \alpha (\sum_{s \in S_0}
w_{s, s'_1} v_1(s) + b_{s'_1}) + (1 - \alpha) (\sum_{s \in S_0}
w_{s, s'_2} v_1(s) + b_{s'_2}) = \sum_{s \in S_0}
(\alpha w_{s, s'_1} + (1- \alpha) w_{s, s'_2}) v_1(s) + (\alpha
b_{s'_1} + (1 - \alpha) b_{s'_2}) = \sum_{s \in S_0}
w_{s, s'_1, s'_2} v_1(s) + b_{s'_1, s'_2})$, where 
$w_{s, s'_1, s'_2} = (\alpha w_{s, s'_1} + (1- \alpha) w_{s, s'_2})$
and $b_{s'_1, s'_2}= \alpha
b_{s'_1} + (1 - \alpha) b_{s'_2}$.
Note that $w_{s, s'_1, s'_2}$ and $b_{s'_1, s'_2}$ are in the edge
weights and biases of the abstract system.
If $\sum_{s \in S_0}
w_{s, s'_2} v_1(s) + b_{s'_2}$ is negative, then $v_2(s'_2) =
v_2(s'_1) = 0$, hence, $\hat{v}_2({\hat{s}}') = 0$ can be simulated
using either the values used to obtain $v_2(s'_1)$ or $v_2(s'_2)$.
If $v_2(s'_1) = 0$, but $v_2(s'_2) > 0$, then we note that $\sum_{s \in S_0}
w_{s, s'_1} v_1(s) + b_{s'_1})  \leq 0$ and $ (\sum_{s \in S_0}
w_{s, s'_2} v_1(s) + b_{s'_2}) > 0$, any linear combination of the two
can still be obtained using $v_1$, and $(1- \alpha) (\sum_{s \in S_0}
w_{s, s'_2} v_1(s) + b_{s'_2})$ is between the two values and can be
obtained from $v_1$, and further, applying $\sigma$ would give us $v_2(s'_2)$. 

\paragraph{Proof of Lemma \ref{lem:main}.} Suppose $(v,v')\in
\dfu{T}{i}$. Then from Proposition \ref{prop:labs}, we have $v'\in
\Post_{\la(\T, i, P_{i+1})}(\ab{v})$, and further, from Proposition
\ref{prop:rabs}, we have $\ab{v'} \subseteq \Post_{\ra(\la(\T, i,
  P_{i+1}),P_i)}(\ab{v})$.
Finally, from Lemma \ref{lem:rl}, we obtain that $\ab{v'} \subseteq
\Post_{\tp, i}(\ab{v})$.

\paragraph{Proof of Theorem \ref{thm:main}.}
Suppose $(v, v') \in \df{T}$, then there exists a sequence of
valuations $v_0, v_1, \cdots, v_k$, where $v_0 = v'$ and $v_i \in
\Post_{\T, i}(v_{i-1})$ for $i > 0$.
From Lemma \ref{lem:main}, we know that since $(v_i, v_{i+1}) \in
\dfu{T}{i}$, $\ab{v_{i+1}} \subseteq
\Post_{\tp, i}(\ab{v_i})$.
Since, $\df{T}$ is the composition of $\dfu{T}{i}$, we obtain that
$\ab{v_k} \subseteq \Post_{\tp}(\ab{v_0})$.
If the nodes in the input and output layer are not merged, then,
$\ab{v_0} = \{v_0\} = \{v\}$ and $\ab{v_k} = \{v_k\} = \{v'\}$.
Therefore, $(v, v') \in \df{T}$.

\paragraph{Proof of Theorem \ref{theorm:encoding}.}
First, let us prove that if $(v, v') \in \dfu{T}{i}$, then  there is a
valuation $z \in \val(Q_i)$, such that $\enc(T, i)$ is satisfied with
values $v, v'$ and $z$.
In fact, it suffices to fix an $s'$ and show that $C_{s'}^{i+1}$ is
satisfied by $v, v'(s')$ and $z(q_{s'})$.
First, note that $v'(s') \geq 0$ since it is obtained by applying the
$\relu$ function, so the second constraint in $C_{s'}^{i+1}$ is
satisfied.
From the semantics, we know that $v'(s') = \sigma(\sum_{s \in S_i}
w_{s, s'} v(s) + b_{s'})$, where $W_i^l(s, s') \leq w_{s, s'} \leq
W_i^u(s, s')$ and $b_i^l(s') \leq b_{s'} \leq b_i^u(s')$.
Hence, $\sigma(\sum_{s \in S_i} W_i^l(s, s') v(s) +b_i^l(s') )  \leq v'(s') \leq \sigma(\sum_{s \in S_i}
W_i^u(s, s') v(s) + b_i^u(s'))$.
Let $v''(s') = \sum_{s \in S_i}
w_{s, s'} v(s) + b_{s'}$, that is, $v'(s') = \sigma(v''(s'))$.

Case $v''(s') \geq 0$: $v'(s') = v''(s')$ and we have $\sum_{s \in
  S_i} W_i^l(s, s') v(s) +b_i^l(s')   \leq v'(s') \leq \sum_{s \in
  S_i} W_i^u(s, s') v(s) + b_i^u(s')$. Hence, for $z(q_{s'}) = 0$, the first,
third and fourth constraints in $C_{s'}^{i+1}$ are satisfied.

Case $v''(s') < 0$: In this case, $v'(s') = 0$ and we set $z(q_{s'}) = 1$.
$\sum_{s \in S_i} W_i^l(s, s') v(s) +b_i^l(s') \leq \sum_{s \in S_i}
w_{s, s'} v(s) + b_{s'} = v''(s') < 0 = v'(s')$, so the first
constraint is satisfied.
$\sum_{s \in S_i} W_i^u(s, s') v(s) + b_i^u(s') + Mq_{s'} = \sum_{s
  \in S_i} W_i^u(s, s') v(s) + b_i^u(s') + M$.
Since, $M$ is an upperbound on the absolute value of $x_{s'}$ before
applying the $\relu$ operation, $\sum_{s
  \in S_i} W_i^u(s, s') v(s) + b_i^u(s') + M$ is positive, and hence,
satisfies the third constraint.
The fourth constraint is satisfied by the choice of $M$, that is, $M
\geq x_{s'}$.

Next, we prove the other direction. Suppose $C_{s'}^{i+1}$ is
satisfied for every $s'$ by some $v, v', z$, then we show that $(v,
v') \in \dfu{T}{i}$.

Case $z(q_{s'}) = 0$: In this case, we have
$\sum_{s \in S_i} \lwr{W_i}(s, s')x_s +\lwr{b_{i}}(s') \leq x_{s'}
\leq  \sum_{s \in S_i} \uppr{W_i}(s, s')x_s +\uppr{b_{i}}(s')$.
Since, $\relu$ is a monotonic function and $x_{s'} > 0$ by the
second constraint, we have $\sigma(x_{s'}) = x_{s'}$ and hence,
$\sigma(\sum_{s \in S_i} \lwr{W_i}(s, s')x_s +\lwr{b_{i}}(s')) \leq
x_{s'} \leq  \sigma(\sum_{s \in S_i} \uppr{W_i}(s, s')x_s
+\uppr{b_{i}}(s'))$.
Hence, $x_{s'} =  \sigma(\sum_{s \in S_i} w_{s, s'} x_s
+ b_{s'})$ for some $\lwr{W_i}(s, s') \leq w_{s, s'} \leq
\uppr{W_i}(s, s')$ and $\lwr{b_{i}}(s')  \leq b_{s'} \leq
\uppr{b_{i}}(s')$.
Hence, $v'(s')$ is obtained from $v$ using the definition of $\dfu{T}{i}$.

Case $z(q_{s'}) = 1$: In this case, $x_{s'} = 0$ and $\sum_{s \in S_i}
\lwr{W_i}(s, s')x_s +\lwr{b_{i}}(s') \leq x_{s'} = 0$.
Therefore $\sigma(\sum_{s \in S_i}
\lwr{W_i}(s, s')x_s +\lwr{b_{i}}(s'))  = 0 = x_{s'}$, therefore
$v'(s')$ is obtained from $v$ using the definition of $\dfu{T}{i}$.